\documentclass[10pt,twocolumn,letterpaper]{article}

\usepackage{wacv}
\usepackage{times}
\usepackage{epsfig}
\usepackage{booktabs}
\usepackage{graphicx}
\usepackage{amsmath}
\usepackage{amssymb}
\usepackage{bm}
\usepackage[inline]{enumitem}
\usepackage{multirow}
\usepackage{xspace}
\usepackage{authblk}
\usepackage{caption}
\usepackage[accsupp]{axessibility}  

\def\wacvPaperID{896} 

\wacvalgorithmstrack   

\wacvfinalcopy 

\ifwacvfinal
    \usepackage[breaklinks=true,bookmarks=false]{hyperref}
\else
    \usepackage[pagebackref=true,breaklinks=true,colorlinks,bookmarks=false]{hyperref}
\fi

\newcommand{\myalgname}{{FreeREA}}
\newcommand{\myalgnamespace}{{FreeREA} } 

\begin{document}
    \title{\myalgname: Training-Free Evolution-based Architecture Search}
    
    \author{
        Niccol\`{o} Cavagnero, Luca Robbiano, Barbara Caputo, Giuseppe Averta\\
        Politecnico di Torino, Italy\\
        {\tt\small\{niccolo.cavagnero, luca.robbiano, barbara.caputo, giuseppe.averta\}@polito.it}
    }
    
    \maketitle
    \thispagestyle{empty}
    
    \begin{abstract}
        In the last decade, most research in Machine Learning contributed to the improvement of existing models, with the aim of increasing the performance of neural networks for the solution of a variety of different tasks. However, such advancements often come at the cost of an increase of model memory and computational requirements. This represents a significant limitation for the deployability of research output in realistic settings, where the cost, the energy consumption, and the complexity of the framework play a crucial role. To solve this issue, the designer should search for models that maximise the performance while limiting its footprint. Typical approaches to reach this goal rely either on manual procedures, which cannot guarantee the optimality of the final design, or upon Neural Architecture Search algorithms to automatise the process, at the expenses of extremely high computational time. This paper provides a solution for the fast identification of a neural network that maximises the model accuracy while preserving size and computational constraints typical of tiny devices. Our approach, named FreeREA, is a custom cell-based evolution NAS algorithm that exploits an optimised combination of training-free metrics to rank architectures during the search, thus without need of model training. Our experiments, carried out on the common benchmarks NAS-Bench-101 and NATS-Bench, demonstrate that i) FreeREA is a fast, efficient, and effective search method for models automatic design; ii) it outperforms State of the Art training-based and training-free techniques in all the datasets and benchmarks considered, and iii) it can easily generalise to constrained scenarios, representing a competitive solution for fast Neural Architecture Search in generic constrained applications. The code is available at \url{https://github.com/NiccoloCavagnero/FreeREA}.
    \end{abstract}
    
    \section{Introduction}
        In recent years, we observed a dramatic expansion of the impact of Machine Learning in several fields, such as education \cite{popenici2017exploring}, industry \cite{lee2018industrial}, and healthcare \cite{rajkomar2019machine}. However, this groundbreaking technological advancement suffers of a major flaw related to the significant impact on the energy consumption and, therefore, to climate change. As an example, a recent study demonstrated how the carbon footprint associated to a single training of common Natural Language Processing models is more than twice the impact of a standard American citizen's daily life for one year \cite{strubell2019energy,dhar2020carbon}. When Neural Architecture Search (NAS) algorithms are used to search for the best model, this amount is increased of a factor ten \cite{strubell2019energy,dhar2020carbon}. 
These observations strongly call for a reduction of the impact that Machine Learning models have in terms of required resources. For this reason, it is likely that current and future trends of research will encompass the development of models that can run on very tiny devices. This represents a significant paradigmatic shift for Machine Learning research. Indeed, when the hardware resources are limited, the designer should pay much more attention on how these resources are used by the model. To do this, Neural Architecture Search \cite{elsken2019neural} appears to be the best option to obtain models that can fully exploit the available budget, with minimum waste of time for human experts.

\begin{figure}[t]
    \centering
    \includegraphics[width=\columnwidth]{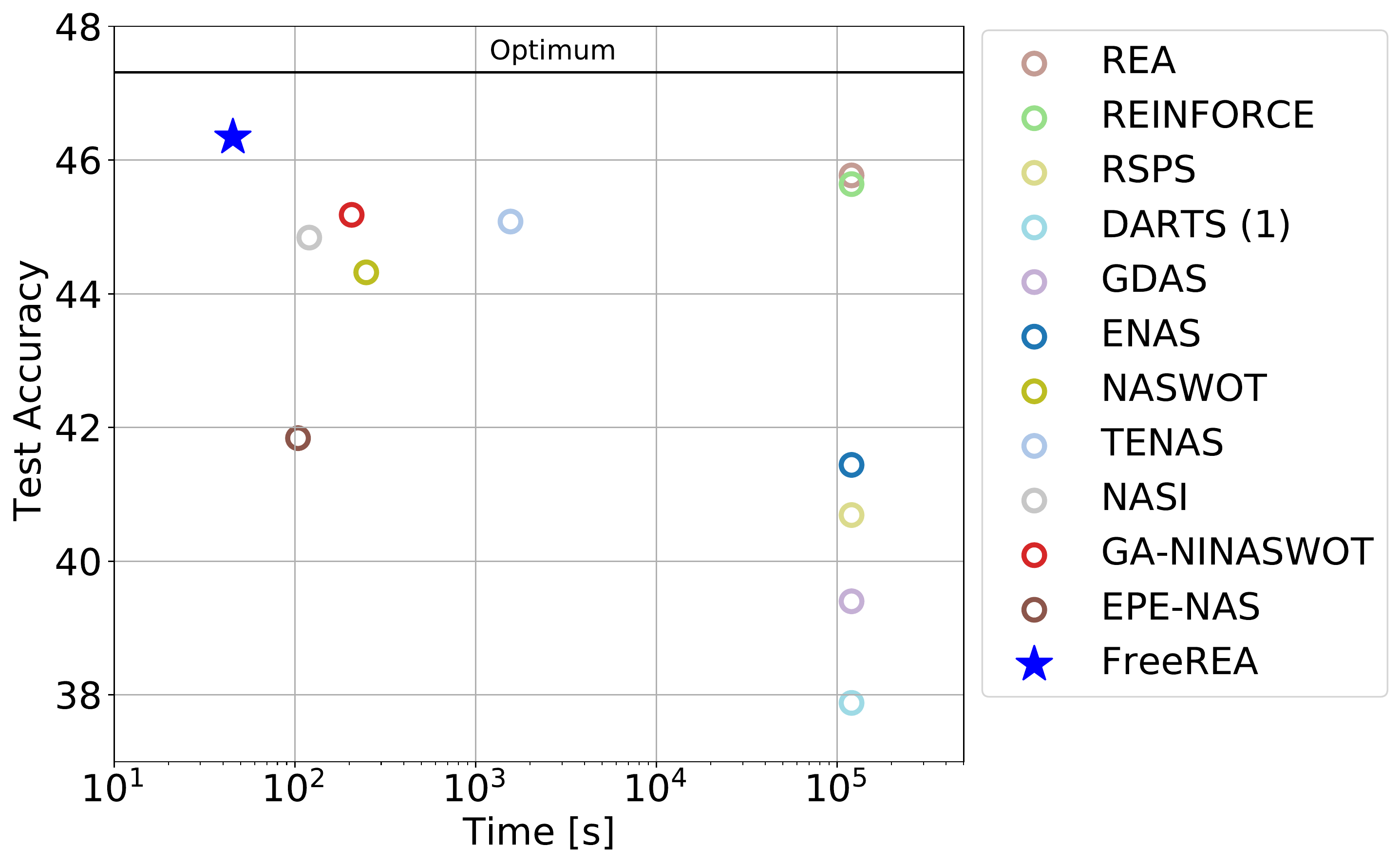}
    \caption{Average test accuracy vs time [s] on ImageNet16-120 of NATS-Bench \cite{nats}. X-axis is in log-scale. Accuracy and times for training-based methods are from the original NATS-Bench paper \cite{nats}. Similar plots can be drawn for the other datasets, and are here omitted for sake of space.}
    \label{fig:acc_vs_time}
    \vspace{-0.5cm}
\end{figure}
However, standard NAS approaches are based on algorithms that - during the search - need to train and test all the candidates to assess their performance. Therefore, NAS pipelines are usually very expensive in terms of computational time and resources, making them hardly exploitable by standard Machine Learning users, with no access to super-computing resources. As an additional drawback, it is worth mentioning that the energy consumption of a standard training-based NAS algorithm would make the effort of designing tiny models vain \cite{strubell2019energy,dhar2020carbon}. 

For these reasons, an effective development of tiny neural networks on large scale requires the implementation of efficient NAS algorithms that do not need train all the candidates. To solve this problem, the research community has recently been focusing on the adoption of proxies to score the architectures at initialisation \cite{naswot,tenas,zero_cost}, thus avoiding the heavy and slow training phase and achieving consistent speedups even with limited hardware resources. 
However, training-free NAS algorithms proposed so far are not competitive with training-based approaches and typically also do not consider the opportunity to include hardware-driven constraints such as model footprint and FLOPs. 

This paper bridges this gap and provides a custom constrained evolutionist method, named \myalgname, to exploit an optimal combination of metrics that serve as proxy of model accuracy. We demonstrate how the metrics we use work as a valid replacement for model training, enabling the possibility to identify automatically highly performing models in just a few minutes of search. We tested our solution on two popular benchmarks for NAS of tiny models, and we demonstrated how our approach outperforms State of the Art methods for the constraint-free case. We also performed, for the first time for a training-free method, experiments in a constrained scenario, setting a very competitive baseline for future works. 

In conclusion, this paper advances the State of the Art with the following contributions:
\begin{enumerate}
    \item an optimised combination of training-free metrics for models ranking, that can completely substitute the training phase in NAS algorithms;
    \item an improved evolutionist search algorithm that fully exploits the ranking strategy to identify very accurate models in few minutes of search;
    \item a number of experiments to demonstrate how training-free methods can effectively replace training-based NAS approaches, even in hardware-constrained scenarios. 
\end{enumerate}
    \section{Related Work}
Neural Architecture Search was firstly introduced in \cite{reinforcement_nas}, where a Reinforcement Learning approach, REINFORCE \cite{reinforce}, with an LSTM \cite{lstm} controller was employed to generate high-performance neural networks. The whole search exploited more than $800$ GPUs for $28$ days, for a total of $22400$ GPU-hours, requiring the partial ($35$ epochs) training of more than 12k different architectures. After this groundbreaking work, the research community focused on the development of more efficient methods \cite{enas,gdas,rea,darts} with the aim of mitigating the huge computational requirements of a NAS algorithm.

The first attempts in this direction focused on Parameter Sharing \cite{enas,gdas,darts}, where newly found architectures inherit their weights from previously discovered ones. While \cite{enas} still adopts a Reinforcement Learning approach, two other search paradigms emerged in NAS literature: differentiable search and evolution. The first ones \cite{darts,gdas} move from the purpose to make the whole search differentiable so that it can be optimised by gradient descent algorithms, leading to significant speedups w.r.t. original methods \cite{reinforcement_nas}. On the other hand, evolution-based strategies are very simple to implement, and naturally allows parameter inheritance from the parents, but suffered of lower performance \cite{evo_nas_review,evo_large}. In REA \cite{rea}, the authors implemented a regularised Tournament Selection search algorithm, which proven to be the first evolution-based NAS able to outperform human-crafted architectures.
Still, these methods were not able to completely solve the core issue, and thousands of different candidates still needed to be trained and evaluated.

For this reason, the research community has recently focused on the adoption of metrics to score the different architectures at initialisation, completely avoiding the training and evaluation phases, which constitute the true bottleneck in NAS algorithms.
The first metric proposed was Linear Regions (NASWOT) \cite{naswot}, which measures the expressivity power of an architecture and allows the search of competitive candidates in few seconds. However, these networks were still less accurate than the ones found by standard NAS algorithms. Therefore, the use of this approach was limited to a mere initialisation for classic search. Later, GA-NINASWOT \cite{ninaswot} adopted Linear Regions together with a genetic algorithm, while EPE-NAS \cite{epe} proposed an improvement over the base metrics. Yet, none of these demonstrated to be competitive with training-based methods. Taking inspiration from NASWOT \cite{naswot}, in \cite{tenas} the authors combined a variation of Linear Regions with the Neural Tangent Kernel (NTK) metric \cite{ntk}, to encode in the search also the trainability of the candidates. Due to the complexity of NTK computation, NASI \cite{nasi} proposed an approximation of the metric obtaining significant speedups. For a detailed analysis on these and other training-free metrics, the interested reader can refer to \cite{zero_cost}.\\
    \section{Method}
        In this paper, we introduce \myalgname, an innovative NAS method that leverages on training-free metrics to deliver high performance architectures satisfying constraints on FLOPs and number of parameters. Indeed, in previous works the training-free phase could not compete with standard search approaches, and was therefore usually exploited only to initialise the population of a classic NAS algorithm \cite{naswot}. In our implementation, instead, we show that a proper selection of metrics, combined with an optimised evolutionist search, can yield accurate models with a significantly lower search time (up to four orders of magnitude lower than standard NAS).
\subsection{Metrics}
\begin{table*}[ht]
    \centering
    \setlength{\tabcolsep}{3pt}
    \caption{Kendall and Spearman correlation between training-free metrics and the test accuracy, evaluated on the three datasets of NATS-Bench \cite{nats}. Each metric has been computed three times with different initialisations and the average is taken as final score.}
    \label{tab:nats_corr}
    \begin{tabular}{ccccccc} 
        \toprule
        \multicolumn{1}{c}{} & \multicolumn{2}{c|}{CIFAR10} & \multicolumn{2}{c|}{CIFAR100} & \multicolumn{2}{c}{ImageNet16-120} \\ 
        \midrule
        \multicolumn{1}{c|}{Metric} & \multicolumn{1}{c|}{Kendall} & \multicolumn{1}{c|}{Spearman} & \multicolumn{1}{c|}{Kendall} & \multicolumn{1}{c|}{Spearman} & \multicolumn{1}{c|}{Kendall} & \multicolumn{1}{c}{Spearman} \\
        
        \midrule
        
        \textbf{NTK} \cite{ntk} & \multicolumn{1}{c}{-0.33} & \multicolumn{1}{c}{-0.49} & \multicolumn{1}{c}{-0.30} & \multicolumn{1}{c}{-0.45} & 
        \multicolumn{1}{c}{-0.39} & \multicolumn{1}{c}{-0.56} \\
        
        \textbf{Snip} \cite{snip} & \multicolumn{1}{c}{0.45} & \multicolumn{1}{c}{0.61} & \multicolumn{1}{c}{0.47} & \multicolumn{1}{c}{0.62} & 
        \multicolumn{1}{c}{0.41} & \multicolumn{1}{c}{0.55} \\
        
        \textbf{Fisher}\cite{fisher} & \multicolumn{1}{c}{0.39} & \multicolumn{1}{c}{0.54} & \multicolumn{1}{c}{0.40} & \multicolumn{1}{c}{0.55} & 
        \multicolumn{1}{c}{0.36} & \multicolumn{1}{c}{0.48} \\
        
        \textbf{Grasp} \cite{grasp} & \multicolumn{1}{c}{0.28} & \multicolumn{1}{c}{0.41} & \multicolumn{1}{c}{0.35} & \multicolumn{1}{c}{0.50} & 
        \multicolumn{1}{c}{0.35} & \multicolumn{1}{c}{0.49} \\
        
        \textbf{PathNorm} \cite{path_norm}& \multicolumn{1}{c}{0.41} & \multicolumn{1}{c}{0.59} & \multicolumn{1}{c}{0.42} & \multicolumn{1}{c}{0.60} & 
        \multicolumn{1}{c}{0.45} & \multicolumn{1}{c}{0.63} \\
        
        \textbf{LinearRegions} \cite{naswot} & \multicolumn{1}{c}{0.61} & \multicolumn{1}{c}{0.79} & \multicolumn{1}{c}{0.62} & \multicolumn{1}{c}{0.81} & 
        \multicolumn{1}{c}{0.60} & \multicolumn{1}{c}{0.78} \\
        
        \textbf{Synflow} \cite{synflow} & \multicolumn{1}{c}{0.57} & \multicolumn{1}{c}{0.77} & \multicolumn{1}{c}{0.56} & \multicolumn{1}{c}{0.76} & 
        \multicolumn{1}{c}{0.56} & \multicolumn{1}{c}{0.75} \\
        
        \textbf{LogSynflow} (ours) & \multicolumn{1}{c}{0.61} & \multicolumn{1}{c}{0.81} & \multicolumn{1}{c}{0.60} & \multicolumn{1}{c}{0.79} & 
        \multicolumn{1}{c}{0.59} & \multicolumn{1}{c}{0.78} \\
        
        \bottomrule
    \end{tabular}
\end{table*}
As anticipated in the previous section, training-free algorithms heavily rely on metrics that serve as accurate proxy of the model performance. Unfortunately, these do not exactly correlate with the test accuracy, and therefore it is important to consider a combination of multiple metrics, possibly accounting for different properties such as trainability and expressivity. With this aim, we performed preliminary experiments on the NATS-Bench dataset to assess the reliability of a selection of metrics as surrogate of the test accuracy. Details on the correlation of different metrics are given in Tab. \ref{tab:nats_corr}. This analysis demonstrates that the most performing metrics are a modified version of Synflow \cite{synflow}, which we named LogSynflow, and the Linear Regions \cite{naswot}. In addition, we complemented these measures with the number of skipped layers to favour more trainable architectures. To combine the different metrics, we sum the normalised scores of the single metrics, which leads to better empirical results w.r.t. a standard cumulative ranking score \cite{tenas}. The rationale behind this choice is that a model that is both trainable and expressive is more likely to show high performances \cite{tenas}. In our implementation, the fitness function $f$ for a given model $i$ can be expressed as:
\begin{equation}
\begin{split}
    f_i = \frac{LS_i}{\max\limits_{j \in J}{LS_j}} + \frac{LR_i}{\max\limits_{j \in J}{LR_j}} + \frac{Skip_i}{\max\limits_{j \in J}Skip_j} \\ 
\end{split}
\end{equation}
where $LS$ stands for LogSynflow, $LR$ for Linear Regions, $Skip$ for Skipped Layers and J is the set of the explored networks. An ablation study on the three terms composing the function can be found in Tab. 1 of the supplementary material.
In addition, to increase the robustness of the measure, we compute both \textit{Linear Regions} and \textit{LogSynflow} three times, with different initialisation, and take the average over the three runs as the final metric. In the following sections we discuss the implementation of the metrics we considered in this work, while we refer the interested reader also to \cite{zero_cost} for further details.

\subsubsection{Linear Regions}
\begin{figure}[t]
    \centering
    \includegraphics[width=0.95 \columnwidth]{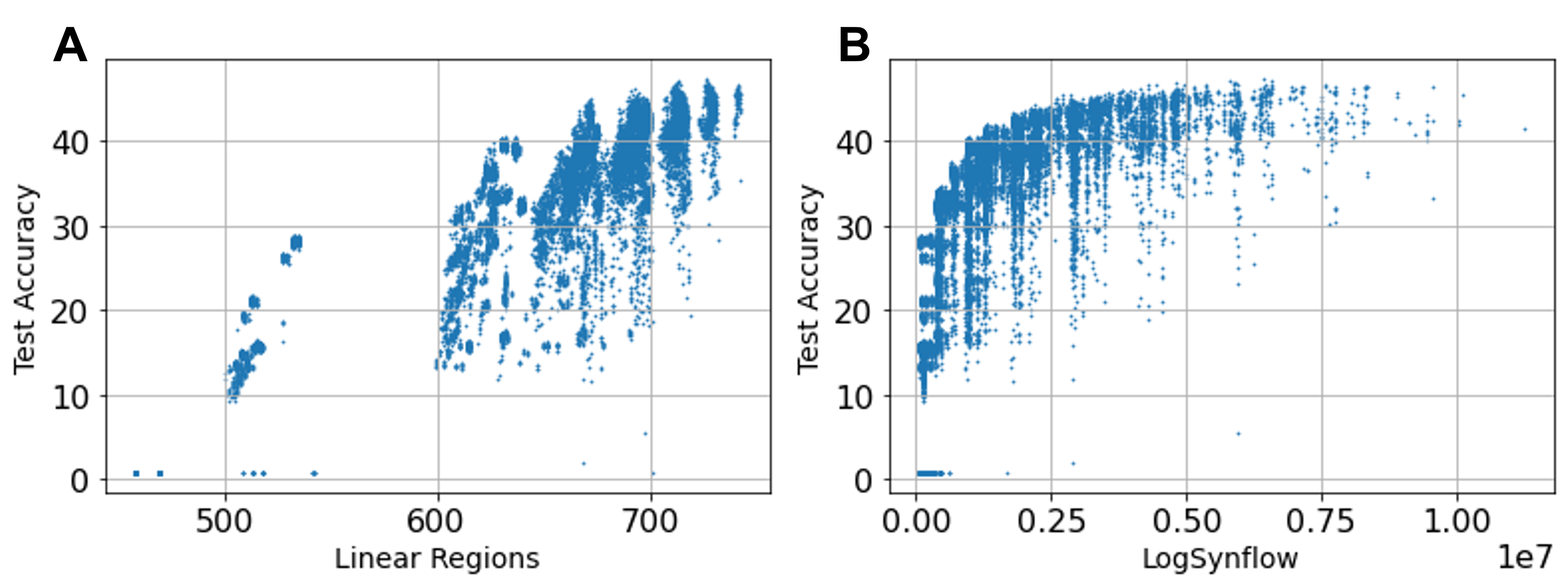}
    \caption{Relationship between test accuracy and the two metrics considered in this work for all the models available in NATS-Bench: Linear Regions in A, \textit{LogSynflow} in B. Accuracy is provided for the ImageNet16-120 dataset. Similar plots can be drawn for the other datasets and are here omitted for sake of space.}
    \label{fig:metrics}
\end{figure}

\textit{Linear Regions} was introduced in \cite{naswot} with the purpose of measuring the expressiveness of an architecture at initialisation. Indeed, when one training sample is forwarded into a ReLU network, the activation values divide the tensors in active (positive values), and inactive (negative values) regions, generating binary masks. Given an input space of fixed dimension, the number of continuous regions of the input space that are mapped to different network activations provides a measure of the model capabilities to discriminate between different input values \cite{naswot}. Indeed, when two different input items result in similar masks, the model cannot easily distinguish between them. As a consequence, an expressive model is capable to map relatively close values in input space to different activation tensors. To quantitatively evaluate this measure, following the implementation of \cite{naswot} we build and forward a mini-batch of data composed of 64 samples into the network, and collect the resulting binary masks. Then, we evaluate the differences in the activation patterns via Hamming distance. Once a Kernel matrix $K_H$ is built from the Hamming distances, the score $s$ is computed as:
\begin{equation}
    s = \log \left\lvert
            \operatorname{det} \left(
                K_H
            \right)
        \right\rvert
\end{equation}
The higher the score, the higher the expressiveness of the network and its ability to distinguish the samples. As depicted in Fig. \ref{fig:metrics}-A, models with high values of Linear Regions score tend to present high test accuracy, as also confirmed by a positive correlation between the two entities (see Tab. \ref{tab:nats_corr}). Therefore, maximising the Linear Regions score yields models that are more expressive and, likely, capable of higher test accuracy.  
\subsubsection{LogSynflow}
\textit{Synflow} was initially introduced in \cite{synflow} as a pruning metric for the selection of weights to be removed to compress an architecture while maximally preserving the performances. Its original implementation locally evaluated the relevance of single weights, and was then extended by \cite{zero_cost} to score a whole architecture by summing up the contributions of the single weights. Formally, the global metric can be defined as the scalar product between the weight vector $\bm{\theta}$ and the gradient vector $\frac{\partial \mathcal{R}}{\partial \bm{\theta}}$:
\begin{equation}
 \mathcal{S}\left(\Theta\right) =
    \bm{\theta}
    \cdot
    \frac {
        \partial \mathcal{R}
        }{
        \partial \bm{\theta}
    }
\end{equation}
To compute the metric, first the network is initialised with the absolute value of its original weights, so that $\theta_i \geq 0$. Then an all-ones tensor is forwarded into the network, and the output is backwarded to compute the gradients. It is worth noticing that, to compute Synflow, Batch Normalisation layers must be suppressed since they interfere with the gradient flow. However, this is likely to result in gradient explosion, even in relatively small architectures. As a consequence, the metric neglects the significance of the weight values because the term associated to the gradients may be several order of magnitudes higher than the corresponding weight. To address this issue, we propose \textit{LogSynflow} which scales down the gradients with a logarithmic function before summing up the contributions of each network weight:
\begin{equation}
  \mathcal{S}\left(\Theta\right) =
    \bm{\theta}
    \cdot
        \log
            \left(\frac {
                \partial \mathcal{R}
                }{
                \partial \bm{\theta}
            } + 1\right)
\end{equation}
To verify that our modification provides a more expressive metric, we computed the numerical values of Synflow and \textit{LogSynflow} over both the benchmark datasets we considered \cite{bench-101,nats} and calculated the correlation of the result with the test accuracy (see Tab. \ref{tab:nats_corr} and \ref{tab:LogSynflow_corr_101}). We observed that our implementation consistently increases the correlation with the test accuracy w.r.t. the standard implementation and, hereinafter we will therefore employ \textit{LogSynflow} in place of the original definition. Furthermore, the positive correlation between test accuracy and \textit{LogSynflow} score, also evident from Fig. \ref{fig:metrics}-B, suggests that searching for high values of \textit{LogSynflow} scores provides architectures that are likely more accurate in test. 

\begin{table}[t]
    \setlength{\tabcolsep}{3pt}
    \caption{Kendall and Spearman correlation between the Synflow/LogSynflow and the test accuracy, evaluated on NAS-Bench-101 \cite{bench-101}. Numerical values suggest that \textit{LogSynflow} is consistently a more accurate proxy of the test accuracy w.r.t. the original definition.}
    \label{tab:LogSynflow_corr_101}
    \centering
    \begin{tabular}{ccc} 
        \toprule
        \multicolumn{1}{c}{Metric} & \multicolumn{1}{c|}{Kendall} & \multicolumn{1}{c}{Spearman} \\
        
        \midrule
        \textbf{Synflow} \cite{synflow} & \multicolumn{1}{c}{0.25} & \multicolumn{1}{c}{0.37} \\
        
        \textbf{LogSynflow} (ours) & \multicolumn{1}{c}{0.31} & \multicolumn{1}{c}{0.45} \\
        \bottomrule
    \end{tabular}
    \vspace{-.5cm}
\end{table}

\subsubsection{\# Skipped Layers}
In recent literature for training-free NAS, Neural Tangent Kernel (NTK) \cite{ntk} is usually employed as a proxy for the trainability of models \cite{tenas,tfmoenas}. However, NTK computation is consistently heavier than other training-free metrics \cite{ntk} and is characterised by a low correlation with the test accuracy (see Tab. \ref{tab:nats_corr}). In order to favour more trainable architectures, we instead adopted as score the number of skipped layers divided by the total number of skip connections in a cell, i.e. $s = \frac{\# \text{Skipped Layers}}{\# \text{Skip Connections}}$. Indeed, since their introduction in \cite{resnet}, which laid the foundation of modern Deep Learning, all the State of the Art networks  (e.g. EfficientNet \cite{efficientnet}, ResNeXt \cite{resnext}, ViT \cite{vit}, and ConvNext \cite{convnext} among the others) present skip connections around convolutional cells in their architecture. These have proven to be crucial for the training of really deep networks and single skip connection around the cell effectively mitigates the vanishing of the gradients and allows the information to be backpropagated more easily through the network. 
The rationale behind this metric is to favour such configuration by encouraging few connections with a large number of skipped layers, i.e. long range connections within a cell. Indeed, the maximum possible value is achieved when there is a single skip connection from input node to output node.

\subsection{Search Algorithm}
To search for the optimal model, we exploited an evolution approach. In particular, an improved Tournament Selection with Ageing (REA) \cite{rea} was implemented. Tournament Selection is initialised with a random population of size $N$. Each individual is characterised by a genotype that encodes the structure of the network. For each iteration, a subset of the surviving population of size $n$ is randomly sampled, and the best individual in the sample is selected as parent for reproduction. The exemplar is then mutated to generate a child which is included into the existing population. Classic Tournament Selection keeps the population size constant by discarding the less fitting individuals, while REA \cite{rea} removes the oldest model, thus biasing the tournament in favour of younger exemplars and therefore enhancing the exploration of the search space. 
With the aim of further improving the search capability of REA, we implemented a variation of the original algorithm which samples two parents at each step. The two parents are independently mutated to generate two children, while a third child is generated through a crossover operation. Crossover is implemented by uniformly sampling a gene from one of the parents. This increases the exploration capability of the search algorithm since, while mutation is inherently local, the combination of two different genotypes can lead to more varied individuals which significantly differ from their parents \cite{evo_nas_review}. We empirically prove the effectiveness of this strategy w.r.t. the base algorithm in the following experiments. To maintain the population size constant, after killing the oldest individual we only keep the top $N$ ones. All gene choices in both mutation and crossover are uniformly sampled.

Constrained search is performed by imposing limits to the FLOPs and to the Number of Parameters for the models considered during the search, because these are not dependent on the specific deployment architecture. In case of constraints, only feasible individuals are added to the population and we keep generating offspring until a feasible child is discovered. In all the experiments of this work, $N$ and $n$ are set to 25 and 5 respectively, if not explicitly reported differently. The results for other values of $N$ and $n$ are reported in Tab. 2 of the supplementary material.
    \section{Benchmarks}
        To help the comparison between our method and State of the Art approaches, and to speed-up the experiments, we leverage on two different benchmarks that include pre-trained models. This enables an easy test of the search algorithms without the need of consistent computational resources, and eliminates the intrinsic variability due to different training pipelines, thus increasing the reproducibility of results. 

\subsection{NAS-Bench-101}
NAS-Bench-101 \cite{bench-101} is the first public dataset built for this purpose. It contains 423k unique convolutional architectures trained and evaluated on CIFAR10. Each network is characterised by a cell, which is repeated with interleaved downsampling operators to form the full structure. Cells are represented by a Direct Acyclic Graph (DAG), with each node corresponding to an operation. There are three possible operators, 1x1 convolution, 3x3 convolution and 3x3 max pooling. The number of nodes in a cell is at most seven and the number of edges is at most nine.

\subsection{NATS-Bench}
NATS-Bench \cite{nats} is an extension of NAS-Bench-201 \cite{nas-bench-201} containing architectures trained and evaluated on three different Computer Vision datasets, CIFAR10, CIFAR100 and ImageNet16-120 (listed from the simplest to the more complex). These are complemented with additional information such as FLOPs and latency, which is particularly relevant for constrained scenarios. However, the number of architectures in this benchmark is limited to 15625, with about 6k unique cells. The possible operators are 1x1 convolution, 3x3 convolution, 3x3 average pooling, skip connection and zeroise. In this case, edges in the DAG represent the operators, while each node stands for the sum of all the feature maps transformed by the edges pointing to the node. The number of edges is six, while the number of nodes is set to four. 
    \section{Experiments}
        Given the lack of experiments testing the efficiency of existing constrained NAS algorithms on the benchmarks mentioned above, we evaluate, as a first step, \myalgnamespace against State of the Art training-based and training-free techniques, without imposing any constraint on both NAS-Bench-101 and NATS-Bench. Then, to also prove its effectiveness in a constrained scenario, we directly compare the architectures discovered by our approach on NATS-Bench against the optimal architectures within the imposed limits. We stop \myalgnamespace after 45 seconds of execution time on NATS-Bench and 12 minutes on NAS-Bench-101. All experiments were performed on a single RTX 3080 GPU. The results for training-based approaches on NATS-Bench are extracted from the original NATS-Bench paper \cite{nats}, while the ones for training-free methods are taken from the respective papers. 
\begin{table*}[ht]
    \setlength{\tabcolsep}{3pt}
    \caption{Test accuracy and time [s] on NATS-Bench. Results for training-based methods are taken from \cite{nats}.}
    \label{tab:unconstrained_nats}
    \centering
    \begin{tabular}{lcccccc}
        \toprule
        \multicolumn{1}{l}{} & \multicolumn{2}{c|}{CIFAR10} & \multicolumn{2}{c|}{CIFAR100} & \multicolumn{2}{c}{ImageNet16-120} \\ 
        \midrule
        \multicolumn{1}{l|}{Algorithm} & \multicolumn{1}{c}{Accuracy} & \multicolumn{1}{c|}{Time} & \multicolumn{1}{c}{Accuracy} & \multicolumn{1}{c|}{Time} & \multicolumn{1}{c}{Accuracy} & \multicolumn{1}{c}{Time} \\ 
        \midrule 
        \multicolumn{7}{c}{\textbf{Training-based}} \\ \midrule
        REA \cite{rea} & $94.02 \pm 0.31$ & 2e4 & $72.23 \pm 0.84 $& 4e4 & $45.77 \pm 0.80$ & 1.2e5 \\
        REINFORCE \cite{reinforcement_nas} & $93.90 \pm 0.26$ & 2e4 & $71.86 \pm 0.89$ & 4e4 & $45.64 \pm 0.78$ & 1.2e5 \\
        RSPS \cite{rsps} & $91.05 \pm 0.66$ & 2e4 & $68.27 \pm 0.72$ & 4e4 & $40.69 \pm 0.36$ & 1.2e5 \\
        DARTS (1) \cite{darts} & $59.84 \pm 7.84$ & 2e4 & $61.26 \pm 4.43$ & 4e4 & $37.88 \pm 2.91$ & 1.2e5 \\
        DARTS (2) \cite{darts} & $65.38 \pm 7.84$ & 2e4 & $60.49 \pm 4.95$ & 4e4 & $36.79 \pm 7.59$ & 1.2e5 \\
        GDAS \cite{gdas} & $93.23 \pm 0.58$ & 2e4 & $68.17 \pm 2.50$ & 4e4 & $39.40 \pm 0.00$ & 1.2e5 \\
        ENAS \cite{enas} & $93.76 \pm 0.00$ & 2e4 & $70.67 \pm 0.62$ & 4e4 & $41.44 \pm 0.00$ & 1.2e5 \\ \midrule 
        \multicolumn{7}{c}{\textbf{Training-free}} \\ 
        \midrule 
        NASWOT (1000) \cite{naswot} & $93.10 \pm 0.31$ & 248 & $69.10 \pm 1.61$ & 248 & $45.08 \pm 1.55$ & 248 \\
        TENAS \cite{tenas} & $93.9 \pm 0.47$ & 1558 & $71.24
        \pm 0.56$ & 1558 & $42.38 \pm 0.46$ & 1558 \\
        NASI \cite{nasi} & $93.55 \pm 0.10$ & 120 & $71.20 \pm 0.14$ & 120 & $44.84 \pm 1.41$ & 120 \\
        GA-NINASWOT \cite{ninaswot} & $93.70 \pm 0.63$ & 206 & $71.57 \pm 1.37$ & 206 & $45.18 \pm 2.05$ & 206 \\
        EPE-NAS \cite{epe} & $91.31 \pm 1.69$ & 104 & $69.58 \pm 0.83$ & 104 & $41.84 \pm 2.06$ & 104 \\
        \myalgname$^-$ (ours) & $94.30 \pm 0.02$ & 45 & $73.30. \pm 0.31$ & 45 & $\textbf{46.34} \pm 0.00$ & 45 \\
        \myalgnamespace (ours) & $\textbf{94.36}\pm 0.00$ & 45 & $\textbf{73.51} \pm 0.05$ & 45 & $\textbf{46.34} \pm 0.00$ & 45 \\ \midrule 
        \multicolumn{7}{c}{\textbf{Optimum}} \\ \midrule 
        \multicolumn{1}{c}{-} & 94.37 & - & 73.51 & - & 47.31 & -
    \end{tabular}
    \vspace{-0.25cm}
\end{table*}
\subsection{Unconstrained Scenario}
\begin{figure}[t]
    \centering
        \includegraphics[width=\columnwidth]{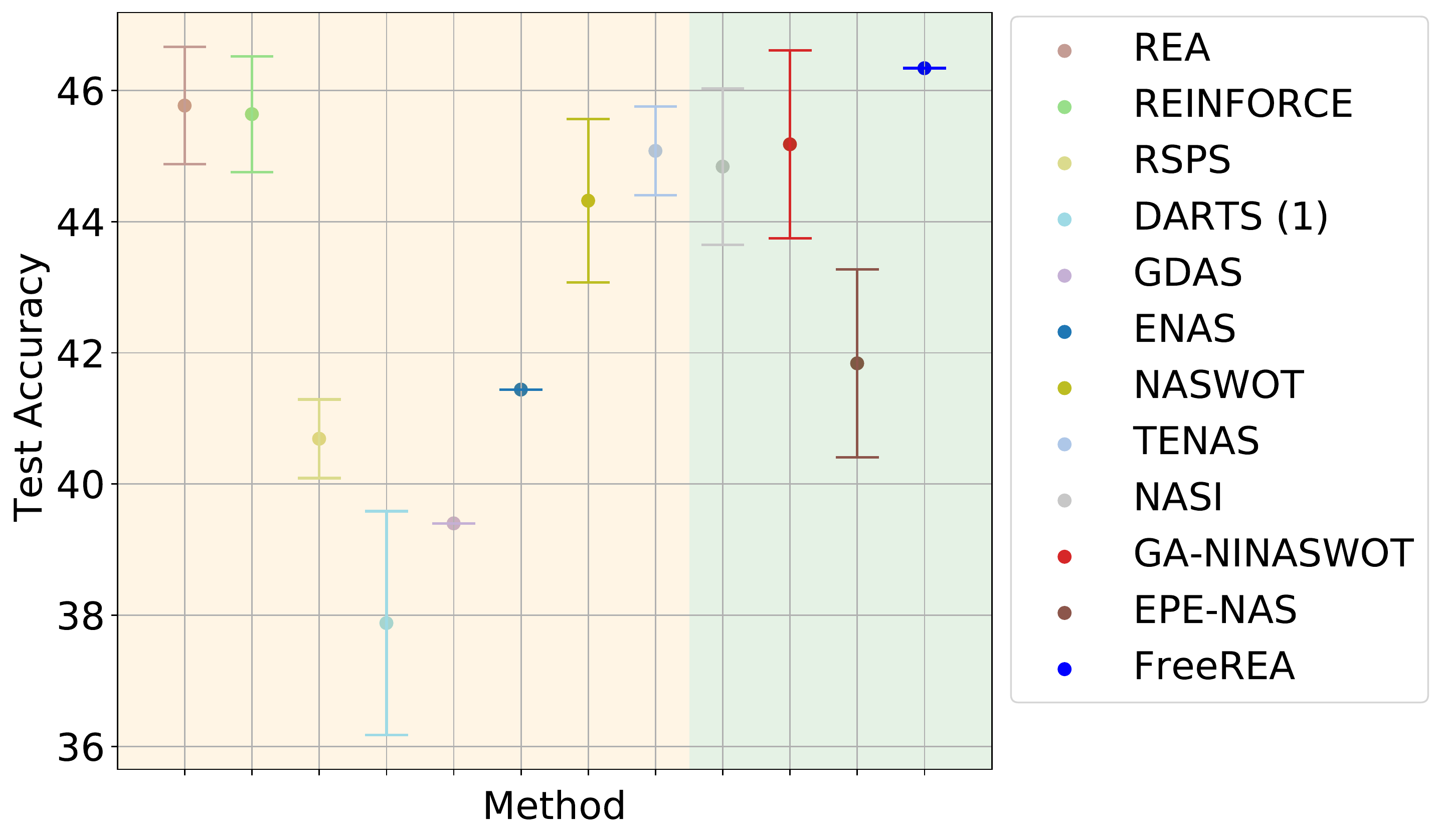}
    \caption{Mean and standard deviation of test accuracy achieved on the ImageNet16-120 dataset of NATS-Bench \cite{nats}. Orange shaded area collects training-based methods, while green shaded area collects training-free methods. Similar plots can be drawn for the other datasets, and are here omitted for sake of space.}
    \label{fig:boxplot}
    \vspace{-.5cm}
\end{figure}
As a first analysis, we considered the NATS-Bench benchmark. As reported in Tab. \ref{tab:unconstrained_nats} and depicted in Fig. \ref{fig:boxplot}, our experiments verified that among the State of the Art approaches REA \cite{rea} is the algorithm that discovers the best architectures. Our solution, however, is able to outperform the previous best result for all the datasets considered in the benchmark, with a gain that increases with the complexity of the task. More specifically, we obtained that \myalgnamespace yields a model with $94.36\%$ of test accuracy for CIFAR10, $73.51\%$ on CIFAR100 and $46.34\%$ for ImageNet16-120, only $0.01\%$, and $0.97\%$ less than the optimum for CIFAR10 and ImageNet16-120, while we hit the optimum in CIFAR100.

Notably, the performance demonstrated by our approach significantly outperforms all the methods (training-based and training-free) considered in this paper, which is especially surprising considering the comparison with training-based approaches. Indeed, while these provide a solution in a (capped) time of 2e4, 4e4 and 1.2e5 seconds, \myalgnamespace iterates for approximately $45$ seconds. 
It is also relevant to notice that our approach is characterised by a significantly smaller variation across different runs (see Tab. \ref{tab:unconstrained_nats} and Fig. \ref{fig:boxplot}). Indeed, the largest variance of the test accuracy we observed (for the CIFAR100 case) is as low as 0.05, while REA shows significantly higher values ($0.31$ for CIFAR-10, $0.84$ for CIFAR100 and $0.80$ for ImageNet16-120). Similar considerations can be made for all the other methods and are here omitted to improve readability of the manuscript. This suggests that our convergence to the optimal model is affected by a smaller uncertainty w.r.t. all the other approaches, thus ideally vouching for a lower number of search runs to obtain the final model. 

It is worth remarking that, while for training-free methods this potentially represents a gain in the order of minutes, for training-based approaches the amount is of several hours or even days. We argue that the increase of average test accuracy and the reduction of variance are a consequence of both the optimal combination of metrics we considered, and of the modifications to the search algorithm we implemented. To verify this claim, we performed experiments in which we used the vanilla REA search algorithm and ranked the models considering our metrics rather than through the validation accuracy after training. We named this implementation \myalgname$^-$. Interestingly, the results achieved with \myalgname$^-$ (see Tab. \ref{tab:unconstrained_nats}) are pretty similar to \myalgname, albeit characterised by a larger variability in certain conditions. Such similarity vouches for the effectiveness of the metrics selected, while it does not provide evidence on the role played by the search algorithm. It is reasonable to expect that this is due to the relative small dimension of the benchmark, while differences should be more evident for benchmarks of higher complexity (such as NAS-Bench-101).

In Fig. \ref{fig:acc_vs_time} we summarise the average test accuracy achieved by our approach and all the competitors versus the execution time. Our findings demonstrate that \myalgnamespace represents the best trade-off between computation time and performance. Of note, all the training-based methods, reported on the right of the plot, are limited by the same search time (see Tab. \ref{tab:unconstrained_nats}). 
To verify the performances of our method on a more challenging scenario, we also tested \myalgnamespace on the more complex search space of NAS-Bench-101 against the other training-free methods and the best-in-class for the training-based (i.e. REA). 
\begin{table}[t]
    \setlength{\tabcolsep}{3pt}
    \centering
    \caption{Test accuracy and time [s] of training-free methods and REA on NAS-Bench-101 \cite{bench-101}.}
    \label{tab:unconstrained_101}
    \begin{tabular}{ccc} 
        \toprule
        \multicolumn{1}{l|}{Algorithm} & \multicolumn{1}{c|}{Accuracy} & \multicolumn{1}{c}{Time} \\ 
        \midrule
        
        & \textbf{Training-based} &\\ 
        \midrule
        
        \multicolumn{1}{l}{REA \cite{rea}} & \multicolumn{1}{c}{$93.39 \pm 0.09$} & \multicolumn{1}{c}{86752} \\ \midrule
        
        & \textbf{Training-free} &\\ 
        \midrule
        
        \multicolumn{1}{l}{NASWOT \cite{naswot}} & \multicolumn{1}{c}{$92.23 \pm 8.90$} & \multicolumn{1}{c}{114} \\ \midrule
        
        \multicolumn{1}{l}{\myalgname$^-$} (ours) & \multicolumn{1}{c}{$93.13 \pm 1.16$} & \multicolumn{1}{c}{723} \\ \midrule
        
        \multicolumn{1}{l}{\myalgname} (ours) & \multicolumn{1}{c}{$\textbf{93.80} \pm 0.02$} & \multicolumn{1}{c}{724} \\ \midrule
        
        & \textbf{Optimum} &\\ 
        \midrule
        \multicolumn{1}{l}{} & \multicolumn{1}{c}{94.31} & \multicolumn{1}{c}{-} \\
        \bottomrule
    \end{tabular}
    \vspace{-0.4cm}
\end{table}
In this experiment, we selected as population and tournament size for REA 50 and 5 respectively. These values are decreased w.r.t. the original implementation provided in NAS-Bench-101 \cite{bench-101} because we allow $\approx24$ hours of search, and only the initialisation of 100 candidates requires $\approx18$ hours. We adopted the same configuration for \myalgname$^-$ for comparison purposes, while the population size is instead kept as default equal to 25 for \myalgname. Numerical results, collected in Tab. \ref{tab:unconstrained_101}, demonstrate how \myalgnamespace still outperforms all the competitors while searching for approximately twelve minutes.

\begin{figure}[t]
    \centering
    \includegraphics[width=\columnwidth]{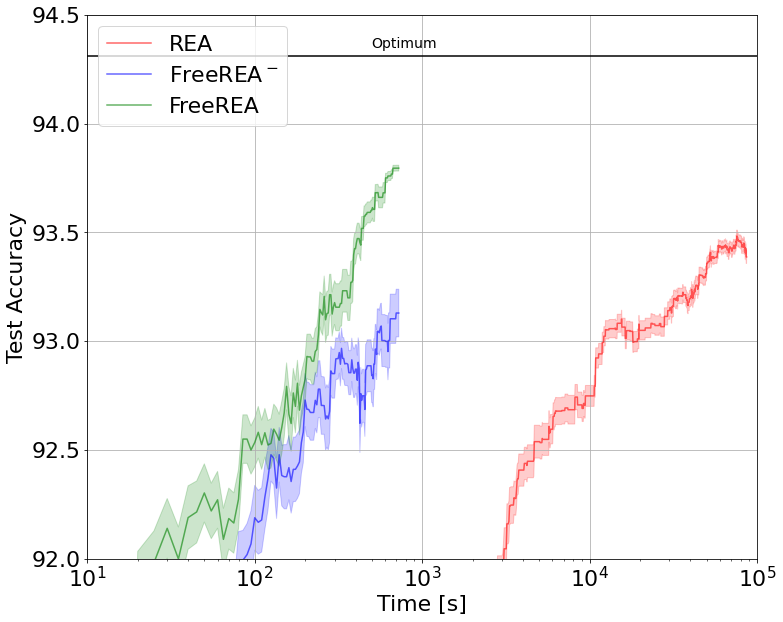}
    \caption{REA, \myalgname$^-$ and \myalgnamespace trajectories on NAS-Bench-101 \cite{bench-101}. Standard deviations are scaled down by a factor 10 for visualisation purposes. X-axis is in log-scale. Note that \myalgnamespace shows a larger variability in the first iterations (i.e. between $10^1$ and $10^2$ s), and a lower uncertainty towards the end of the search. REA and \myalgname$^-$, instead, present similar variances in all the phases of the search.}
    \label{fig:reas}
    \vspace{-0.5cm} 
\end{figure}

Interestingly, there is an appreciable difference between \myalgnamespace and \myalgname$^-$ in terms of average test accuracy, which supports our claim that the search algorithm of our implementation is more efficient than the vanilla REA. Indeed, the modifications we introduced, discussed in the previous section, play a fundamental role in giving better exploration capabilities to the search algorithm. In other words, by mutating two parents we better inspect the neighbourhood of highly performing models, while by means of crossover we also explore more distant regions of the search space. As a consequence, we expect that the difference in terms of test accuracy between \myalgnamespace and \myalgname$^-$ increases with the complexity of the search space and the exploration time. To prove this, we collected the average test accuracy and its variance for \myalgname, \myalgname$^-$ and REA versus the computation time (see Fig. \ref{fig:reas}). It is also interesting to observe that \myalgnamespace shows a larger variability in the first iterations (i.e. between $10^1$ and $10^2$ s), while the output shows a significantly smaller uncertainty towards the end of the search. This behaviour is not matched by \myalgname$^-$ and REA, and is a consequence of the modifications made to the search algorithm, which suggests that our implementation not only provides higher exploration capabilities, but also lower search uncertainty. 

\subsection{Constrained Scenario}
As we discussed extensively in the introduction, it is our opinion that future trends of ML research will devote an increasing effort towards the development of architectures that can fit on constrained devices. In this scenario, albeit the role of models optimisation is fundamental to maximise the exploitation of the available hardware, we believe that there should not be explicit modifications to the search algorithm that depend on the target hardware. For this reason, we tested our approach in a constrained scenario without explicitly accounting for the target constraints, besides the introduction of limits to the exploration phase. 

It is worth reporting that the introduction of those constraints is not straightforward, and good performances of the method in an unconstrained scenario do not necessarily correspond to analogous results in the constrained case. Indeed, \myalgnamespace heavily relies on metrics that serve as proxy of the test accuracy, but the correlation between the two in the whole search space may be different than the correlation in a constrained one. Therefore, an efficient training-free NAS algorithm should be able to search for a good model with and without limits imposed by the target hardware, ideally with no change to the metrics and the search algorithm used. 
\begin{table}[t]
    \setlength{\tabcolsep}{3pt}
    \centering
    \caption{Test accuracy of \myalgnamespace on NATS-Bench in constrained scenarios. Regret is the difference between optimum and average test accuracy. The algorithm was stopped after 45 seconds.}
    \label{tab:constrained}
    \begin{tabular}{cccccc} 
        \toprule
        \multicolumn{1}{c|}{FLOPs} &
        \multicolumn{1}{c|}{Params} & \multicolumn{1}{c|}{Accuracy} & \multicolumn{1}{c|}{Optimum} &
        \multicolumn{1}{c}{Regret}\\ 
        \midrule
        \multicolumn{5}{c}{\textbf{CIFAR10}} \\ 
        \midrule
        
        \multicolumn{1}{c}{1e8} & \multicolumn{1}{c}{8e5} & \multicolumn{1}{c}{$94.12 \pm 0.01$} & \multicolumn{1}{c}{94.31} & \multicolumn{1}{c}{0.19} \\
        
        \multicolumn{1}{c}{7e7} & \multicolumn{1}{c}{5e5} & \multicolumn{1}{c}{$93.02 \pm 0.13$} & \multicolumn{1}{c}{93.75} & \multicolumn{1}{c}{0.73} \\
        
        \multicolumn{1}{c}{4e7} & \multicolumn{1}{c}{3e5} & \multicolumn{1}{c}{$91.30 \pm 0.00$} & \multicolumn{1}{c}{91.32} & \multicolumn{1}{c}{0.02} \\
        
        \midrule
        \multicolumn{5}{c}{\textbf{CIFAR100}} \\ 
        \midrule
        
        \multicolumn{1}{c}{1e8} & \multicolumn{1}{c}{8e5} & \multicolumn{1}{c}{$72.08 \pm 0.01$} & \multicolumn{1}{c}{72.43} & \multicolumn{1}{c}{0.35} \\
        
        \multicolumn{1}{c}{7e7} & \multicolumn{1}{c}{5e5} & \multicolumn{1}{c}{$71.07 \pm 0.00$} & \multicolumn{1}{c}{71.63} & \multicolumn{1}{c}{0.54} \\
        
        \multicolumn{1}{c}{4e7} & \multicolumn{1}{c}{3e5} & \multicolumn{1}{c}{$68.39 \pm 0.00$} & \multicolumn{1}{c}{68.48} & \multicolumn{1}{c}{0.09} \\
        
        \midrule
        \multicolumn{5}{c}{\textbf{ImageNet16-120}} \\ 
        \midrule
        
        \multicolumn{1}{c}{1e8} & \multicolumn{1}{c}{8e5} & \multicolumn{1}{c}{$46.08 \pm 0.08$} & \multicolumn{1}{c}{46.85} & \multicolumn{1}{c}{0.77} \\
        
        \multicolumn{1}{c}{7e7} & \multicolumn{1}{c}{5e5} & \multicolumn{1}{c}{$45.35 \pm 0.04$} & \multicolumn{1}{c}{45.95} & \multicolumn{1}{c}{0.65} \\
        
        \multicolumn{1}{c}{4e7} & \multicolumn{1}{c}{3e5} & \multicolumn{1}{c}{$40.73 \pm 0.00$} & \multicolumn{1}{c}{41.00} & \multicolumn{1}{c}{0.27} \\
        
        \bottomrule
    \end{tabular}
    \vspace{-0.5cm} 
\end{table}
Since, to the best of our knowledge, there are no experimental results in constrained scenarios for the considered benchmarks, to verify whether our implementation represents a good solution also in this case, we considered three different thresholds for model FLOPs and number of parameters, namely (1e8, 8e5), (7e7, 5e5), and (4e7, 3e5), which stand for three levels of hardware constraints. We then tested \myalgnamespace on the three datasets of NATS-Bench for all the constraints. 

In absence of other competitors, in Tab. \ref{tab:constrained} we compare the results of our search with the best model available in the dataset that fits the constraints, which clearly represents an upper-bound for our problem. Interestingly, on the CIFAR datasets the average margin of improvement left (regret) is $0.32\%$, while on the more challenging ImageNet16-120, we under-perform the optimal model of $0.56\%$ on average. 

It is worth remarking that such results are coherent with the unconstrained scenario, where for the more challenging ImageNet16-120 the difference between the test accuracy of our best candidate and the optimal model is on average approximately $0.97\%$. Therefore, this suggests that our combination of metrics serves as a good surrogate of the test accuracy even when very hard constraints are imposed to the search, and that these results may serve as a strong baseline for future research on this topic. 
    \section{Conclusions}
        In this paper, we presented \myalgname, a new fast, efficient and accurate training-free NAS for tiny models. Moving from State of the Art results, we selected a suitable combination of training-free metrics that serve as an accurate proxy of the model test accuracy, and we used this index to rank different networks during the search. The latter is implemented following an evolution-based policy, yielding very competitive results in a fraction of the time required by competitor approaches. We tested our method on two different benchmarks, NATS-Bench \cite{nats} (with three datasets and $\approx15k$ available models as search space), and NAS-Bench-101 \cite{bench-101} (with one dataset and $\approx500k$ available models as search space). Interestingly, our results demonstrate that \myalgnamespace not only is able to effectively perform Neural Architecture Search without training any candidate, but it also is the first training-free method which competes and outperforms State of the Art training-based methods, resulting in the current most efficient and accurate NAS algorithm on the considered benchmarks. Such advancement comes also with the relevant benefit that our search time is as low as twelve minutes in the worst case, i.e. up to four order of magnitude lower than State of the Art algorithms \cite{rea}. Additional experiments on constrained settings suggest that \myalgnamespace can deliver accurate models even when the target hardware imposes limits to the search space that exclude some of the models. Motivated by these results, our future efforts will be devoted towards the generalisation of our method to other search spaces, and considering more challenging tasks. 
    
    {\small
        \bibliographystyle{ieee_fullname}
        \bibliography{egbib}
    }

\end{document}


\title{\myalgname: Training-Free Evolution-based Architecture Search \\ Supplementary}
\author{
    Niccol\`{o} Cavagnero, Luca Robbiano, Barbara Caputo, Giuseppe Averta\\
    Politecnico di Torino, Italy\\
    {\tt\small\{niccolo.cavagnero, luca.robbiano, barbara.caputo, giuseppe.averta\}@polito.it}
    }
\def\wacvPaperID{896}
\maketitle

\section{Fitting Function Ablation}
With the aim of assessing the contribution of different terms composing our fitting function, we performed an ablation study by removing each one of the three components (leave one out). The different configurations are tested on all the datasets from the two benchmarks taken into account, NATS-Bench \cite{nats} and NAS-Bench-101 \cite{bench-101}. We list in Table \ref{tab:fit_ablation} the results achieved by our method with a subset of metrics.
\begin{table*}[!h]
\centering
\caption{Ablation Study for different metrics composing the fitting function. Test Accuracy for the final model is reported. Each configuration has been run 30 times with a time limit of 45 seconds on NATS-Bench and 12 minutes on NAS-Bench-101.}
\label{tab:fit_ablation}
\begin{tabular}{lcccc}
\toprule
 & \multicolumn{3}{c|}{NATS-Bench} & \multicolumn{1}{c}{NAS-Bench-101} \\ \midrule
 & \multicolumn{1}{c|}{CIFAR10} & \multicolumn{1}{c|}{CIFAR100} & \multicolumn{1}{c|}{ImageNet16-120} & \multicolumn{1}{c}{CIFAR10} \\ \midrule
\multicolumn{1}{l|}{Baseline} & \textbf{94.36 $\pm$ 0.00} & \textbf{73.51 $\pm$ 0.00} & \textbf{46.34 $\pm$ 0.00} & \textbf{93.80 $\pm$ 0.02} \\ \midrule
\multicolumn{1}{l|}{w/o  Linear Regions} & \textbf{94.36 $\pm$ 0.00} & \textbf{73.51 $\pm$ 0.00} & \textbf{46.34 $\pm$ 0.00} & 93.55 $\pm$ 0.49 \\ \midrule
\multicolumn{1}{l|}{w/o Skipped Layers} & 93.76 $\pm$ 0.00 & 71.11 $\pm$ 0.00 & 41.44 $\pm$ 0.00 & 91.90 $\pm$ 1.69 \\ \midrule
\multicolumn{1}{l|}{w/o  LogSynflow} & 94.30 $\pm$ 0.00 & 71.13 $\pm$ 0.00 & 44.48 $\pm$ 0.00 & 92.62 $\pm$ 0.08 \\ \bottomrule
\end{tabular}
\end{table*}

Interestingly, only the combination of all the three metrics demonstrated to achieve the highest test accuracy with the lowest variance on all the considered benchmarks. Remarkably the metrics we proposed, Skipped Layers and LogSynflow, seem to be the most important among the three, while the contribution of Linear Regions is limited and appreciable on NAS-Bench-101 only, probably because its discrimination is more effective on larger search spaces.

\section{Search Algorithm Hyper-parameters Ablation}
In addition, we also investigated different choices for $N$ and $n$, population size and tournament size, following the couples explored in the original REA \cite{rea}. Table \ref{tab:hyper_ablation} reports on the experimental results.
\begin{table*}[t]
\centering
\caption{Ablation Study for different choices of $N$ and $n$ on NATS-Bench \cite{nats}. Test Accuracy for the final model is reported. Each configuration has been run 30 times with a time limit of 45 seconds.}
\label{tab:hyper_ablation}
\begin{tabular}{c|ccc}
\toprule
\multicolumn{1}{c|}{N, n} & \multicolumn{1}{c|}{CIFAR10} & \multicolumn{1}{c|}{CIFAR100} & ImageNet16-120 \\ \midrule
25, 5 & \textbf{94.36 $\pm$ 0.00} & \textbf{73.51 $\pm$ 0.00} & \textbf{46.34 $\pm$ 0.00} \\ 
100, 2 & 93.95 $\pm$ 0.14 & 71.8 $\pm$ 1.85 & 45.65 $\pm$ 2.57 \\
100, 50 & \textbf{94.36 $\pm$ 0.00} & \textbf{73.51 $\pm$ 0.00} & \textbf{46.34 $\pm$ 0.00} \\
20, 20 & \textbf{94.36 $\pm$ 0.00} & \textbf{73.51 $\pm$ 0.00} & \textbf{46.34 $\pm$ 0.00} \\
100, 25 & 94.34 $\pm$ 0.01 & 73.50 $\pm$ 0.00 & \textbf{46.34 $\pm$ 0.00} \\
64, 16 & \textbf{94.36 $\pm$ 0.00} & \textbf{73.51 $\pm$ 0.00} & \textbf{46.34 $\pm$ 0.00} \\ \bottomrule
\end{tabular}
\end{table*}
Numerical values suggest that almost all the configurations yield similar performances, vouching for an high robustness of the proposed method over the choices of the hyper-parameters. Interestingly, only the configuration (100, 2) has achieved a significantly lower test accuracy, likely caused by the large difference between $N$ and $n$. It is also worth reporting that, since we allow the generation of offspring from the top two candidates in the sample, with $n=2$ Tournament Selection is not happening, while the algorithm is simply evolving random networks in the population. These two factors, while improving the exploration of the search space, severely hinder the exploitation capability of the algorithm leading to poor results. 

As a rule of thumb, then, it is reasonable to set a lower bound for $\frac{n}{N}$ such that $\frac{n}{N}\geq 0.20$. It is worth remarking that higher value of $n$ and $N$ may result in higher generalisation and exploration capabilities of the algorithm, but may slow down the search. However, in our experiments all the configurations are run for $45$ seconds and proved to converge to the same optimal model, thus suggesting that a longer convergence time is not affecting the performance of the method for the NATS-Bench search space.

{\small
\bibliographystyle{ieee_fullname}
\bibliography{egbib}
}